\documentclass[11pt]{article}
\usepackage{chicagor}
\usepackage{times}

\newtheorem{THEOREM}{Theorem}[section]
\newenvironment{theorem}{\begin{THEOREM}}%
                        {\end{THEOREM}}
\newtheorem{LEMMA}[THEOREM]{Lemma}
\newenvironment{lemma}{\begin{LEMMA}}%
                      {\end{LEMMA}}
\newtheorem{COROLLARY}[THEOREM]{Corollary}
\newenvironment{corollary}{\begin{COROLLARY}}%
                          {\end{COROLLARY}}
\newtheorem{PROPOSITION}[THEOREM]{Proposition}
\newenvironment{proposition}{\begin{PROPOSITION}}%
                            {\end{PROPOSITION}}
\newtheorem{DEFINITION}[THEOREM]{Definition}
\newenvironment{definition}{\begin{DEFINITION}}%
                            {\end{DEFINITION}}
\newtheorem{CLAIM}[THEOREM]{Claim}
\newenvironment{claim}{\begin{CLAIM}}%
                            {\end{CLAIM}}
\newtheorem{EXAMPLE}[THEOREM]{Example}
\newenvironment{example}{\begin{EXAMPLE}}%
                            {\end{EXAMPLE}}
\newtheorem{REMARK}[THEOREM]{Remark}
\newenvironment{remark}{\begin{REMARK}}%
                            {\end{REMARK}}

\newcommand{\thm}{\begin{theorem}}
\newcommand{\lem}{\begin{lemma}}
\newcommand{\pro}{\begin{proposition}}
\newcommand{\dfn}{\begin{definition}}
\newcommand{\rem}{\begin{remark}}
\newcommand{\xam}{\begin{example}}
\newcommand{\cor}{\begin{corollary}}
\newcommand{\prf}{\noindent{\bf Proof:} }
\newcommand{\ethm}{\end{theorem}}
\newcommand{\elem}{\end{lemma}}
\newcommand{\epro}{\end{proposition}}
\newcommand{\edfn}{\bbox\end{definition}}
\newcommand{\erem}{\bbox\end{remark}}
\newcommand{\exam}{\bbox\end{example}}
\newcommand{\ecor}{\end{corollary}}
\newcommand{\eprf}{\bbox\vspace{0.1in}}
\newcommand{\beqn}{\begin{equation}}
\newcommand{\eeqn}{\end{equation}}

\newcommand{\bbox}{\vrule height7pt width4pt depth1pt}

\newcommand{\clm}{\begin{claim}}
\newcommand{\eclm}{\end{claim}}
\newcommand{\commentout}[1]{}
\renewcommand{\phi}{\varphi}

\renewcommand{\gets}{=}
\renewcommand{\L}{{\cal L}}
\renewcommand{\S}{{\cal S}}
\newcommand{\U}{{\cal U}}
\newcommand{\V}{{\cal V}}
\newcommand{\R}{{\cal R}}
\newcommand{\T}{{\cal T}}
\newcommand{\F}{{\cal F}}

\newcommand{\W}{\Omega}
\newcommand{\true}{{\it true}}
\newcommand{\false}{{\it false}}
\newcommand{\union}{\cup}
\newcommand{\inter}{\cap}
\newcommand\eqdef{=_{\rm def}}
\newcommand{\infers}{\vdash}
\newcommand{\dimp}{\Leftrightarrow}
\newcommand{\rimp}{\Rightarrow}
\newcommand{\sat}{\models}
\newcommand{\<}{\langle}
\renewcommand{\>}{\rangle}

\newcommand{\AX}{\mbox{AX}}
\newcommand{\AXp}{\mbox{AX}'}

\newcommand{\AXun}{\mbox{AX}_{\rm uniq}}

\newcommand{\Trec}{{\cal T}_{\rm rec}}
\newcommand{\M}{{\cal M}}
\newcommand{\Mrec}{{\cal M}_{\rm rec}}
\newcommand{\Tun}{{\cal T}_{\rm uniq}}
\newcommand{\Lprop}{{\cal L}_{\rm uniq}}
\newcommand{\LGP}{{\cal L}_{\rm GP}}
\newcommand{\Lex}{{\cal L}^+}

\newcommand{\closest}{\mathit{closest}}
\newcommand{\shortv}{\commentout}
\newcommand{\journal}{\commentout}
\newcommand{\fullv}[1]{#1}
\newcommand{\ML}{\mathit{MD}}
\newcommand{\vdashp}{\vdash^+}
\newcommand{\FF}{\mathit{FF}}

\begin{document}

\title{From Causal Models To Counterfactual Structures\thanks{A
preliminary version of this paper appears in the 
Proceedings of the Twelfth International
Conference on  Principles of Knowledge Representation and Reasoning (KR
2010), 2010.}}
\author{
Joseph Y. Halpern\thanks{
Supported in part by NSF grants IIS-0534064, IIS-0812045, and
IIS-0911036, and by AFOSR grants 
FA9550-08-1-0438 and FA9550-09-1-0266, and ARO grant W911NF-09-1-0281.}\\
Computer Science Department\\
Cornell University\\
halpern@cs.cornell.edu}
\maketitle 
\newcommand{\RCond}{\succeq}   

\begin{abstract}
Galles and Pearl \citeyear{GallesPearl98} claimed that ``for recursive
models, the causal model 
framework does not add any restrictions to counterfactuals, beyond those
imposed by Lewis's [possible-worlds] framework.''   This claim is
examined carefully, with the goal of clarifying the exact relationship
between causal models and Lewis's framework.   Recursive models are shown to
correspond precisely to a 
subclass of (possible-world) counterfactual structures.  On the other
hand, a slight generalization of recursive models, models where all
equations have unique solutions, is 
shown to be incomparable in expressive power to counterfactual
structures, despite the fact that the Galles and Pearl arguments should
apply to them as well.  The problem with the Galles and Pearl argument
is identified: an axiom that they viewed as irrelevant, because it
involved disjunction (which was not in their language), is not
irrelevant at all.  
\end{abstract}

\section{Introduction}

Counterfactual reasoning arises in broad array of fields, from statistics
to economics to law.  Not surprisingly, there has been a great deal of
work on giving semantics to counterfactuals.  Perhaps the best-known
approach is due to Lewis \citeyear{Lewis73} and Stalnaker
\citeyear{Stalnaker68}, and involves possible worlds.  The idea is that
a counterfactual of the form ``if $A$ were the case then $B$ would be
the case'', typically written $A \RCond B$, is true at a world $w$ if
$B$ is true at all the worlds closest to $w$ where $A$ is true.  Of
course, making this precise requires having some notion of ``closeness''
among worlds.

More recently, Pearl \citeyear{pearl:2k} proposed the use of causal
models based on \emph{structural equations} for reasoning about
causality.  In causal models, we can examine 
the effect of \emph{interventions}, and answer questions of the form
``if random variable $X$ were set to $x$, what would the value of random
variable $Y$ be''.  This suggests that causal models can also provide
semantics for (at least some) counterfactuals.  

The relationship between the semantics of counterfactuals 
in causal models and in counterfactual structures (i.e., possible-worlds
structures where the semantics of counterfactuals is given in terms of
closest worlds) was studied by
Galles and Pearl \citeyear{GallesPearl98}.  They argue
that the 
relationship between the two approaches
depends in part on whether we consider
\emph{recursive} 
(i.e., \emph{acyclic}) models (those without feedback---see
Section~\ref{sec:review} for details).  They reach the following conclusion
\cite[p. 242]{pearl:2k}.%
\footnote{The discussion in Section~7.4.2 of \cite{pearl:2k} is taken
almost verbatim from \cite{GallesPearl98}; since the former is more
widely available, that is what I cite in this paper.}
\begin{quote}
In sum, for recursive models, the causal model framework does not add
any restrictions to counterfactuals  beyond those imposed by Lewis's
framework; the very general concept of closest worlds is sufficient.  Put
another 
way, the assumption of recursiveness is so
strong that it already embodies all other restrictions imposed by
structural semantics.  When we consider nonrecursive systems, however,
we see that reversibility [a particular axiom introduced by 
Galles and Pearl] is not enforced by Lewis's framework.
\end{quote}
This conclusion seems to have been accepted by the community.  For
example, in the Wikipedia article on ``Counterfactual conditional''
(en.wikipedia.org/wiki/Counterfactual\_conditional; Sept., 2009), it says 
``[The definition of counterfactuals in causal models] 
has been shown to be compatible with the axioms of
possible world semantics.''

In this paper I examine these claims and the proofs given for them more
carefully, and try to settle once and for all the relationship between
causal models and counterfactual structures.
The first sentence in the statement above
says ``for recursive models, the causal model
framework does not add any restrictions to counterfactuals  beyond those
imposed by Lewis's framework''.  It is not clear (at least to me)
exactly  what this means.  Recursive models are a well-defined subclass
of causal models.  Galles and Pearl themselves show that there are
additional axioms that hold in recursive models over and above those
that hold in counterfactual structures. Indeed, they show that the
reversibility axiom mentioned 
above is valid in recursive models and is not valid in possible-worlds
models.  They also show that all the axioms 
of causal reasoning in the possible-worlds framework that they view as
relevant (specifically, axioms that do not mention disjunction, since it
is not in their language) hold in recursive causal models.  Thus, the only 
conclusion that can be drawn from their argument is just the
opposite to what they claim, namely, that the possible-worlds approach does
not add any restrictions to causal reasoning beyond those imposed by
causal models, since causal models satisfy all the axioms that
counterfactual structures do, and perhaps more.%
\footnote{Although it is not relevant to the focus of this paper, I in
fact also do not
understand the second sentence in the Pearl quote above.  As for
the third sentence, while it is the case that 
reversibility does not hold in counterfactual structures in general, 
reversibility holds in recursive causal structures as well as more general
causal structures, as Pearl and Galles themselves show.}   

Pearl [private communication, 2010] intended the clause ``for recursive
models'' to apply to counterfactual structures as well as to structural
models.  However, the notion of a recursive counterfactual structure is not
defined either by Galles and Pearl \citeyear{GallesPearl98} or Pearl
\citeyear{pearl:2k}.  In fact, in general, the notion of recursive model as
defined, for example, in \cite[Definition 7.3.4]{pearl:2k}, does not
even make sense for counterfactual structures.  I show that, if we focus on
the language considered by Galles and Pearl and counterfactual
structures appropriate for this language, there is a well-defined
subclass of counterfactual structures that can justifiably be viewed as
recursive (counterfactual) structures.  I then show that precisely the
same formulas in this language are valid in recursive causal models and
recursive 
counterfactual structures.   Put another way, at least as far as Galles
and Pearl's language goes, recursive causal models and recursive
counterfactual structures are equally expressive.  
Thus, by restricting to recursive counterfactual structures (as Pearl
intended), the Galles-Pearl claim is, in a sense, correct (although the
claim does not follow from their argument---see below).  However, it should be
noted that Galles and Pearl's language cannot, for example, express
statements with disjunctive antecedents, such as ``if he had chosen a
different running mate or had spent his campaign funding more wisely
then he would have won the election''.  The claim applies only to their
restricted language.

Galles and Pearl try to prove their claims by considering axioms; my
proof works at the level of structures.  Specifically, 
I show that a recursive causal model satisfying a
particular formula can be effectively converted to a recursive
counterfactual structures satisfying this formula, and vice versa. 
It is actually important to work at the level of structures to prove
this claim, rather than working at the level of axioms.  
I show that the argument that Galles and Pearl give for
their claim is incorrect in some significant respects.  The problem 
lies with their claim that axioms involving disjunctive antecedents are
irrelevant. 
As I show, they are quite relevant; a proof of a formula not 
involving disjunction in the antecedent may need to use an axiom that does.  
This possibility is illustrated by considering a class of
models slightly larger than recursive models.
In a recursive model, given a context, there is a unique
solution to every equation.  In earlier work \cite{Hal20}, I showed that
there are nonrecursive causal models where every equation has a unique
solution.  Galles and Pearl's argument applies without change to causal
models where the equations have a unique solution.  However, as I show
here, these models are actually \emph{incomparable} in expressive
power to
counterfactual structures.  The reversibility axiom remains valid in
causal models where the equations have a unique solution.  But, as I show
by example, there is a formula that does not involve disjunction that is valid
in counterfactual structures but
is \emph{not} valid in causal models where the equations have
a unique solution.  Not surprisingly, proving that this formula is valid
from the axioms requires the use of an axiom that involves disjunction.

These results show that the quote from Wikipedia is not quite true.
While it is true that \emph{recursive} causal structures are, in a
certain sense,
compatible with the axioms of possible worlds semantics, the slightly more
general class of causal structures with unique solutions to the
equations is not.  Thus, in general,  the semantics of counterfactuals
in causal structures \emph{cannot} be understood in terms of closest
worlds.  

\fullv{
The rest of this paper is organized as follows. In
Section~\ref{sec:review}, I review the causal models and counterfactual
structures.  The main technical results are in Section~\ref{sec:main},
which also includes a discussion of the problems with the Galles-Pearl
argument.   I conclude in Section~\ref{sec:discussion} with some
discussion of these results.
}

\fullv{\section{Causal models and possible-worlds models: a
review}\label{sec:review}  }
\shortv{\section{Causal models and possible-worlds models}\label{sec:review}} 

In this section I review the relevant material on causal models and
possible-worlds models.

\subsection{Causal models}\label{sec:causalmodel}
(The following discussion is taken, with minor modifications, from
\cite{Hal20}.)    
Causal models describe the world in terms of random variables,
some of which have a causal effect on others.  
It is conceptually useful to split the random variables into two
sets, the {\em exogenous\/} variables, whose values are determined by
factors outside the model, and the
{\em endogenous\/} variables, whose values are ultimately determined by
the exogenous variables.   The values of the endogenous variables are
characterized by a set of {\em structural equations}.

For example, if we are trying to determine
whether a forest fire was caused by lightning or an arsonist, we 
can take the world to be described by four random variables:
\begin{itemize}
\item $\FF$ for forest fire, where $\FF=1$ if there is a forest fire and
$\FF=0$ otherwise; 
\item $L$ for lightning, where $L=1$ if lightning occurred and $L=0$ otherwise;
\item $\ML$ for match dropped (by arsonist), where $\ML=1$ if the arsonist
dropped a lit match, and $\ML = 0$ otherwise;
\item $E$, which captures the external factors that determine whether
the arsonist will drop a match, or whether lightning will strike.
\end{itemize}
The variables $\FF$, $L$, and $\ML$ are endogenous, while $E$ is
exogenous.
If we want to model the fact that
if the arsonist drops a match \emph{or} lightning strikes then a fire
starts, then we would have the equation
$\FF = \max(L,\ML)$; that is, the value of the random variable $\FF$ is the
maximum of the values of the random variables $\ML$ and $L$.  This 
equation says, among other things, that if $\ML=0$ and $L=1$, then
$\FF=1$.
Alternatively, if we want to model the fact that a fire requires both a
lightning strike \emph{and} a dropped match (perhaps the wood is so wet
that it needs two sources of fire to get going), then the only change in the
model is that the equation for $\FF$ becomes $\FF = \min(L,\ML)$; the
value of $\FF$ is the minimum of the values of $\ML$ and $L$.  The only
way that $\FF = 1$ is if both $L=1$ and $\ML=1$.  

Formally, a {\em signature\/} $\S$ is a tuple $(\U,\V,\R)$,
where $\U$ is a finite set of exogenous variables, $\V$ is a finite set
of endogenous variables, and $\R$ associates with every variable $X \in
\U \union \V$ a finite set $\R(X)$
of possible values for $X$ (the {\em range\/} of possible values of
$X$).  
A {\em causal model\/} is a pair $T=(\S,\F)$, where $\S$ is a signature and
$\F$ associates with each variable $X \in \V$ a function denoted
$F_X$ such that $F_X: (\times_{Z \in (\U \union \V - \{X\})} \R(Z))
\rightarrow \R(X)$. 
$F_X$ characterizes the value of $X$
given the values of all the other variables in $\U \union \V$.
Because $F_X$ is a function, there is a unique value of $X$ once all the
other variables are set. 
Notice that we have such functions only for the endogenous variables.
The exogenous variables
are taken as given; it is their effect on the endogenous
variables (and the effect of the endogenous variables on each other)
that is modeled by the structural equations.

Given a causal model $T = (\S,\F)$, a (possibly empty)  vector
$\vec{X}$ of variables in $\V$, and a vector $\vec{x}$
of values for the variables in
$\vec{X}$, we can define a new causal model
denoted
$T_{\vec{X} \gets \vec{x}}$. 
Intuitively, this is the causal model that results when the variables in
$\vec{X}$ are set to $\vec{x}$.
We can think of setting $\vec{X}$ to $\vec{x}$ as an intervention.  For
example, if $T$ is the causal model for the forest fire described above,
where $\FF = \max(L,\ML)$, 
then $T_{L=0}$ is the model where the
lightning definitely does not occur, so that there is a forest fire if
and only if the arsonist drops a match.  If $T'$ is the model where $\FF
= \min(L,\ML)$, then $T'_{L=0}$ is the model where there is no forest
fire, since there is no lightning. 

Formally, $T_{\vec{X} \gets \vec{x}} = (\S,
\F^{\vec{X} \gets \vec{x}}\})$,
where $\F^{\vec{X} \gets \vec{x}}$ is identical to $\F$ except that the
equation for $X$ is replaced by the equation $X = x$.
The model  $T_{\vec{X} \gets \vec{x}}$ describes a possible {\em
counterfactual\/} situation; that is, even though, under normal
circumstances, setting the exogenous variables to $\vec{u}$ results
in the variables $\vec{X}$ having value $\vec{x}' \ne \vec{x}$, this
submodel describes what happens if they are set to $\vec{x}$ due to
some ``external action'', the cause of which is not modeled explicitly.

In general, given a \emph{context} $\vec{u}$, that is, a setting for the
exogenous variables, there may not be a unique vector of values that 
simultaneously satisfies all the equations in $T_{\vec{X} \gets
\vec{x}}$; indeed, there may not be a solution at all.  One
special case where the equations in a causal model $T$ are guaranteed to
have a unique solution is 
when there is a total ordering $\prec_T$ of the variables in $\V$ such
that if $X \prec_T Y$, then $F_X$ is independent of the value of $Y$;
that is,
$F_X(\ldots, y, \ldots) = F_X(\ldots, y', \ldots)$ for all $y, y' \in
\R(Y)$.  In this case, $T$ is said to be {\em
recursive\/} or {\em acyclic\/}.  Intuitively, if $T$
is recursive, then there is no
feedback; if $X \prec_T Y$, then the value of $X$ may affect the value of
$Y$, but the value of $Y$ has no effect on the value of $X$.
It should be clear that if $T$ is a recursive model, then, given a
context $\vec{u}$, there is
always a unique solution to the equations in
$T_{\vec{X} \gets \vec{x}}$, for all $\vec{X}$ and $\vec{x}$,
(We simply solve for the variables in the order given by
$\prec_T$.)  

Following \cite{Hal20}, I consider three successively more
general classes of causal models for a given signature $\S$ (with a
focus on the first two):
\begin{enumerate}
\item $\Trec(\S)$: the class of recursive causal models over
signature $\S$;
\item $\Tun(\S)$: the class of causal models $T$ over $\S$
where, for all $\vec{X} \subseteq \V$, $\vec{x}$,
the equations in $T_{\vec{X} \gets \vec{x}}$ 
have a unique solution for all contexts $\vec{u}$;
\item $\T(\S)$: the class of all causal models over $\S$.
\end{enumerate}
I often omit the signature $\S$ when it is clear from
context or irrelevant, but the reader should bear in mind its
important role.  

\paragraph{Syntax and Semantics:}  In \cite{Hal20}, a number of
languages for reasoning about causality are considered.  
The choice of language is significant.  As Galles and Pearl
already point out, we cannot in any obvious way give a meaning in causal
models to counterfactual implications where there is a disjunction on
the left-hand side of the implication, that is formulas of the form $(A
\lor A') \RCond B$.  Thus, our results effectively consider a language
with no disjunction on the left-hand side of $\RCond$.  As mentioned in
the introduction, one of the results 
of this paper have the form ``for every recursive causal model, there is
a recursive causal structure that satisfies the same formulas'', and
vice versa.  For this result to have any bite, we must choose a
reasonably rich language.

In \cite{Hal20}, I considered a number of languages appropriate for
reasoning about causality in causal models.  I briefly review two of them here.
The languages are
parameterized by the signature $\S$.  A {\em basic causal formula\/} is
one of the form $[Y_1 \gets y_1, \ldots, Y_k \gets y_k] \phi$, where
$\phi$ is a Boolean combination of formulas of the form $X = x$, $Y_1,
\ldots, Y_k, X$ are variables in $\V$, $Y_1, \ldots, Y_k$ are
distinct, and $x \in \R(X)$.  I typically abbreviate such a formula
as $[\vec{Y} \gets \vec{y}]\phi$.
The special case where $k=0$ is abbreviated as $\phi$.
A {\em causal formula\/} is a Boolean combination of basic causal formulas.
Let $\Lex(\S)$ consist of all causal formulas over the signature $\S$.  
(Again, I often omit the signature $\S$ if it is clear from context or
not relevant.)  

Roughly speaking, we can think of $\Lex(\S)$ as the language that
results by starting with primitive propositions of the form $X=x$, 
where $X$ is a random variable in $\V$ and $x \in \R(X)$, and
closing under modal operators of the form $[\vec{Y} \gets \vec{y}]$.
The restriction to primitive propositions of this form is not a major one.
Given a propositional language with primitive propositions
$P_1, \ldots, P_n$, we can define binary random variables $X_1, \ldots,
X_n$ (i.e., variables whose range is $\{0,1\}$) and identify $X_i = 1$
with ``$P_i$ is true''.  That is, as long as we can we define structural
equations that characterize how a change in one primitive propositions
affects the other primitive propositions, taking the primitive
propositions to have the form $X=x$ is not a major restriction.  

The formula $[\vec{Y} \gets \vec{y}] X = x$ can be interpreted as ``in all
possible solutions to the structural equations obtained after setting
$Y_i$ to $y_i$, $i = 1, \ldots, k$, and the exogenous variables to
$\vec{u}$, random variable $X$ has value $x$''. 
This formula is true in a causal model
$T$ in a context $\vec{u}$ if in all solutions to the equations in
$T_{\vec{Y} \gets \vec{y}}$ in context $\vec{u}$, the random variable
$X$ has value $x$.
Note that this formula is trivially true if there are no
solutions to the structural equations.

\commentout{
Just as with dynamic logic, we
can also define the formula $\<\vec{Y} \gets
\vec{y}\>(X(\vec{u}) = x)$ to be an abbreviation of
$\neg [\vec{Y} \gets \vec{y}] \neg (X(\vec{u}) = x)$.
$\<\vec{Y} \gets \vec{y}\> (X(\vec{u}) = x)$ is the dual of
$[\vec{Y} \gets \vec{y}] (X(\vec{u}) = x)$; it is true if, in some
solution to the structural equations obtained after setting $Y_i$ to
$y_i$, $i = 1, \ldots, k$, and the exogenous variables to $\vec{u}$,
random variable $X$ has value $x$.
 Taking $\true(\vec{u})$ to be an abbreviation for
$X(\vec{u})
= x
\lor
X(\vec{u}) \ne x$ for some variable $X$ and $x \in \R(X)$, and taking
$\false(\vec{u})$ to be an abbreviation for $\neg \true(\vec{u})$, we
have that $\<\vec{Y}
\gets \vec{y}\>\true(\vec{u})$ is true if there is some solution to the
equations obtained by setting $Y_i$ to $y_i$, $i = 1, \ldots, k$, and
the variables in $\U$ to $\vec{u}$
(since $[\vec{Y} \gets \vec{y}]\false(\vec{u})$ says that in every
solution to the equations obtained by setting $Y_i$ to $y_i$ and $\U$ to
$\vec{u}$, the formula
$\false(\vec{u})$ is true, and thus holds exactly if the equations
have no solution).
}

\commentout{
Following Galles and Pearl's notation, I often write
$[\vec{Y} \gets \vec{y}]X(\vec{u}) = x$ as
$X_{\vec{Y} \gets \vec{y}}(\vec{u})
= x$. If $\vec{Y}$ is clear from context or
irrelevant, I further
abbreviate this as $X_{\vec{y}}(\vec{u}) = x$.  (This is actually the
notation used by Galles and Pearl.)
Let $\LGP(\S)$ be the sublanguage of $\Lprop(\S)$ consisting of just
conjunctions of formulas
of the form $[\vec{Y} \gets \vec{y}](X = x)$.  In particular, it does
not contain disjunctions or negations of such formulas.  Although Galles and
Pearl \citeyear{GallesPearl98} are not explicit about the
language they are using, it seems to be $\LGP$.
}

A formula in $\Lex(\S)$ is true or false in a
causal model in $\T(\S)$, given a context $\vec{u}$.  As usual,
we write $(T,\vec{u}) \sat \phi$ if the causal formula $\phi$ is true in
causal model $T$ given context $\vec{u}$.%
\footnote{In \cite{Hal20}, following \cite{GallesPearl98}, the context
was included in the formula.  The definition of $(T,\vec{u}) \sat X=x$
given here is intended to be equivalent to that of $T \sat X(\vec{u}) =
x$.  The advantage of having $\vec{u}$ on the left-hand side of $\sat$
(which is also the formalism used in \cite{HP01b}) is that it enforces
the intuition that the context 
consists of background information that is typically suppressed. 
It does lead to loss in expressive power, since it is not possible to consider
formulas of the form $X_1(\vec{u}_1) = x_1 \land X_2(\vec{u}_2) = x_2$,
where $\vec{u}_1 \ne \vec{u}_2$.  But this turns out
to be an advantage---see the next footnote.}
For a basic causal formula $[\vec{Y} \gets \vec{y}]\phi$, we define
$(T,\vec{u}) \sat [\vec{Y} \gets \vec{y}]\phi$ if $\phi$ holds in all
solutions to the equations $\F^{\vec{Y} \gets \vec{y}}$ with the values
of the variables in $\U$ set to $\vec{u}$.  Thus, for example, if $\phi$ 
has the form $X_1 = x_1 \lor X_2 = x_2$, then $(T,\vec{u}) \sat \phi$
iff  every vector
of values for the endogenous variables that simultaneously satisfies all
the equations in $\F$ has either $X_1 =x_1$ or $X_2 = x_2$.%
\footnote{In \cite{Hal20}, truth was defined for formulas of the
form $[\vec{Y} \gets \vec{y}](X=x)$, and extended in ``the obvious way''
to Boolean combinations; that is $\phi_1 \lor \phi_2$ was taken to be
true if either $\phi_1$ was true or $\phi_2$ was true.  This is
equivalent to the approach described above as long as all equations have
a unique solution, that is, in $\Tun$.    
However, the two approaches are not equivalent if equations can have more
than one solution.  The approach suggested here is what is required to
make the axioms given in \cite{Hal20} sound in the general case.  
We do not want to say that $X_1 = x_1 \lor X_2 = x_2$ is true if $X_1 =
x_1$ is true in all solutions to the equations or $X_2 = x_2$ is true in
all solutions to the equations; rather, we want either $X_1 = x_1$ or
$X_2 = x_2$ to be true in all solutions.
With the approach given above, it is not clear how to give semantics to
formulas such as  $X_1(\vec{u}_1) = x_1 \lor X_2(\vec{u}_2) = x_2$ if
$\vec{u}_1 \ne \vec{u}_2$, which is perhaps another good reason for not
allowing such formulas.}
We define the truth value of arbitrary causal formulas, which
are just Boolean combinations of basic causal formulas, in the obvious
way:
\begin{itemize}
\item $(T,\vec{u}) \sat \phi_1 \land \phi_2$ if $(T,\vec{u}) \sat \phi_1$ and $(T,\vec{u}) \sat
\phi_2$
\item $(T,\vec{u}) \sat  \neg\phi$ if $(T,\vec{u}) \not\sat \phi$.
\end{itemize}
As usual, a formula $\phi$ is said to be {\em valid\/} with respect to
a class $\T'$ of causal models if $(T,\vec{u}) \sat \phi$ for all $T \in
\T'$ and contexts $\vec{u}$ in $T$.

$\Lex$ is the most general language that I consider in \cite{Hal20}.  To
compare my results to those of GP, who use a more restricted language, I
also consider some restrictions of $\Lex$.  
Specifically, let $\Lprop$ be the sublanguage of $\Lex$
that consists of Boolean combinations of
formulas of the form $[\vec{Y} \gets \vec{y}](X = x)$.
Thus, the difference between $\Lprop$ and
$\Lex$ is that in $\Lprop$, only $X = x$ is allowed after
$[\vec{Y} \gets \vec{y}]$, while in $\Lex$, arbitrary Boolean
combinations of formulas of the form $X = x$ are allowed.
As the following lemma, proved in \cite{Hal20}, shows, for reasoning
about causality in $\Tun$, the language 
$\Lprop$ is adequate, since it is equivalent in expressive power to
$\Lex$. 

\lem\label{lem:equiv} {\rm \cite{Hal20}} In $\Tun$ and $\Trec$, the language
$\Lex$ and $\Lprop$ are expressively equivalent; for every formula $\phi
\in \Lex$, we can effectively find a formula $\phi' \in \Lprop$ such
that $\Tun \sat \phi \dimp \phi'$.
\elem

The equivalence described in Lemma~\ref{lem:equiv} no longer holds when
reasoning about 
causality in the more general class $\T$ of structures.  However, since
I focus in this paper on $\Trec$ and $\Tun$, in the rest of the paper
I consider the language $\Lprop$; this suffices to make my points.

\paragraph{Axiomatizations}
I briefly recall some standard definitions from logic.  An {\em axiom
system\/}\index{axiom
system}~AX consists of a collection of {\em axioms\/}\index{axiom|(}
and {\em inference rules}\index{inference rule|(}.
An axiom is a formula (in some predetermined language ${\cal L}$), and
an inference
rule has the form ``from $\phi_1, \ldots, \phi_k$ infer~$\psi$'',
where $\phi_1, \ldots, \phi_k, \psi$ are formulas in ${\cal L}$.
A {\em proof\/} in AX consists of a
sequence of formulas in ${\cal L}$, each of which is either an axiom
in~AX or follows by an application of an inference rule.
A proof is said to be a {\em proof of the formula~$\phi$} if the last
formula in the proof is~$\phi$.
We say~$\phi$ is {\em provable in~AX},
and write $\mbox{AX} \infers \phi$,
if there is a proof of~$\phi$ in~AX; similarly, we say that $\phi$ is
{\em consistent with AX\/} if $\neg \phi$ is not provable in AX.

An axiom system AX is said to be {\em sound\/} for
a language ${\cal L}$ with respect to a class
$\T'$ of causal models if every formula in ${\cal L}$
provable in AX is valid with respect to $\T'$.
AX is {\em complete\/} for
${\cal L}$ with respect to
${\cal T'}$ if every formula in ${\cal L}$
that is valid with respect to $\T'$ is
provable in AX.

Consider the following axioms, taken from \cite{Hal20}, modified
slightly for the language used here:
\begin{list}{}{\setlength{\labelwidth}{2in}\setlength{\leftmargin}{.3in}} 
\item[C0.] All instances of propositional tautologies in the language
$\Lprop$. 
\item[C1.] $[\vec{Y}\gets \vec{y}] (X = x) \rimp \neg[\vec{Y} \gets \vec{y}]
(X = x')$, if $x, x' \in \R(X), x \ne x'$.
\hfill  (Equality)
\item[C2.] $\lor_{x \in \R(X)}[\vec{Y} \gets \vec{y}]  (X= x)$.
\hfill (Definiteness)
\item[C3.] $([\vec{X} \gets \vec{x}](W = w) \land ([\vec{X} \gets
\vec{x}](Y = y)) \rimp [\vec{X} \gets \vec{x}; W \gets w](Y=y)$).
\footnote{Galles and Pearl use a stronger version of C3: $([\vec{X} \gets
\vec{x}](W = w) \rimp ([\vec{X} \gets  
\vec{x}](Y = y)) \dimp [\vec{X} \gets \vec{x}; W \gets w](Y=y))$.  The
stronger version follows from the weaker version in the presence of the
other axioms.  (This is actually shown in the proof of
Proposition~\ref{pro:recursive}.)} 
\hfill (Composition)
\item[C4.] $[X \gets x; \vec{W} \gets \vec{w}] (X = x)$. 
\hfill (Effectiveness)
\item[C5.] $([\vec{X} \gets \vec{x}; W \gets w](Y = y) 
\land  [\vec{X} \gets \vec{x}; Y \gets y](W = w)) \\
\mbox{}\ \ \ \rimp [\vec{X} \gets \vec{x}](Y=y)$, \  if $Y \ne W$.%
\footnote{The assumption that $Y \ne W$ was not made by Galles and Pearl
\citeyear{GallesPearl98} nor by Halpern \citeyear{Hal20}, but it is
necessary.  For example, if $W=Y$, then one instance of C5 would be 
$([\vec{X} \gets \vec{x}; Y \gets y](Y = y) \land 
[\vec{X} \gets \vec{x}; Y \gets y](Y = y)) 
\rimp [\vec{X} \gets \vec{x}](Y=y)$.  The antecedent is true by C4,
while the conclusion is not true in general.}
\hfill (Reversibility)
\end{list}
The key axioms C3--C5 were introduced (and named) by Galles and Pearl
\citeyear{GallesPearl98}.  Perhaps most relevant to this paper is the
reversibility axiom, C5.  It
says that
if setting $\vec{X}$ to $\vec{x}$ and $W$ to $w$ results in $Y$ having
value $y$ and setting $\vec{X}$ to $\vec{x}$ and $Y$ to $y$ results in
$W$ having value $w$, then $Y$ must already have value $y$ when we set
$\vec{X}$ to $x$ (and $W$ must already have value $w$).  

Let $\AXun(\S)$ consist of C0--C5 and the rule of inference \emph{modus
ponens} (from $\phi$ and $\phi \rimp \psi$ infer $\psi$).  

\thm\label{thm:ax} ({\rm \cite{Hal20}})
$\AXun(\S)$ is a sound and complete
axiomatization for $\Lprop(\S)$ with respect to $\Tun(\S)$.
\ethm

Using Lemma~\ref{lem:equiv}, it is possible to get a complete
axiomatization for $\Lex$ with respect to $\Tun(\S)$, simply by adding
axioms for converting a formula in $\Lex$ to an equivalent formula in
$\Lprop$.  A complete axiomatization for $\Lprop$ with respect to
$\Trec(\S)$ is also given in \cite{Hal20}; it requires adding a somewhat
complicated axiom called C6 to $\AXun$ that captures acyclicity.  The
details are not relevant for our purposes,   so I do not discuss C6 further.

Finally, a complete axiomatization
for $\Lex$ with respect to $\T(\S)$, the class of all causal models, is
given.  The axioms are similar in spirit to those in $\AXun$.  In
particular, there is the following analogue to reversibility (where $\<
\vec{X} \gets \vec{x} \> \phi$ is an abbreviation for $\neg [\vec{X}
\gets \vec{x}] \neg \phi$):
$$\begin{array}{ll}
(\<\vec{X}\gets \vec{x}; Y \gets y\> (W = w \land
\vec{Z} = \vec{z})  \land
\<\vec{X}\gets \vec{x}; W \gets w\> (Y = y \land
\vec{Z} = \vec{z}))\\
\mbox{$ \ \ \ $}\rimp \<\vec{X} \gets \vec{x}\> (W
= w \land Y = y \land \vec{Z} = \vec{z})),
\mbox{$\ \ \ \ $  where $\vec{Z} = \V - (\vec{X} \union \{W,Y\})$.}
\end{array}$$

\subsection{Possible-worlds models for counterfactuals}

There have been a number of semantics for counterfactuals.  I focus
here on one due to Lewis \citeyear{Lewis73}.  
Let $\Phi$ be a finite set of primitive propositions.  A {\em counterfactual
structure\/} $M$ is a tuple $(\W,R,\pi)$, where $\W$ is a finite set of
{\em possible  worlds\/},%
\footnote{Giving semantics to counterfactual formulas in structures with
infinitely many worlds adds an extra level of complexity.  As shown 
by Friedman and Halpern \citeyear{FrH3}, if a formula in the language I
am about to introduce is 
satisfiable at all, it is satisfiable in a structure with finitely many
worlds, so as far as validity goes, there is no loss of generality in
restricting to finite structures.}  
$\pi$ is an \emph{interpretation} that maps each possible world to a truth
assignment over $\Phi$, and $R$ is a ternary relation over $\W$. 
Intuitively, $(w,u,v) \in R$ if $u$ is as
close/preferred/plausible as $v$ when the real world is $w$. 
Let $u \preceq_w v$ be an abbreviation for $(w,u,v) \in R$, and define
$\W_w = \{u : u \preceq_{w} v \in R 
\mbox{ for some $v \in \W$}\}$; thus, the
worlds in $\W_w$ are those that are at least as plausible as some
 world in $\W$ according to $\preceq_w$.  
Define $v \prec_w v'$ if $v
 \preceq_w v'$ and $v' \not\preceq_w v$.  
We require that $w \in \W_w$, that 
$\preceq_w$ be reflexive and transitive on $\W_w$, and that
$w \prec_w u$ for all $u \ne w$.
Thus, $\prec_w$ puts an ordering on worlds that can be viewed as
characterizing ``closeness to $w$'',
and $w$ is the closest world to itself.

Let $\L^C(\Phi)$ be the language formed by starting with $\Phi$
and closing off under $\land$, $\neg$, and $\RCond$ (where $\RCond$ denotes
counterfactual implication).  The language $\L^C$ allows arbitrary
nesting of counterfactual implications.   By way of contrast, the
language $\Lex$ and its sublanguages have only one level of nesting, if
we think of $[\vec{X} = \vec{x}]\phi$ as $(\vec{X} = \vec{x}) \RCond
\phi$.  Let $\L^C_1$ be the sublanguage of $\L^C$ consisting of all
formulas with no nested occurrence of $\RCond$ (including formulas with
no occurrence of $\RCond$ at all).

We can give semantics to formulas in $\L^C$ (and hence $\L^C_1$) in
a counterfactual structures $M = (\W,R,\pi)$ as follows.  The first few
clauses are standard:
\begin{itemize}
 \item $(M,w) \models p$, when $p \in \Phi$, if $\pi(w)(p) = {\bf true}$.
 \item $(M,w) \models \phi \land \psi$ if $(M,w) \models \phi$ and
    $(M,w) \models \psi$.
 \item $(M,w) \models \neg\phi$ if it is not the case that $(M,w)
    \models \phi$.
\end{itemize}
To give semantics to $\phi \RCond \psi$, assume inductively that we have
already given semantics to $\phi$ at all worlds in $M$.  Define
$\closest_M(w,\phi) = \{v \in \W_w: (M,v) \sat \phi$ and there is no world $v'$
such that $(M,v') \sat \phi$ and $v' \prec_w v\}$.  Thus,
$\closest_M(w,\phi)$ is the set of worlds closest to $w$ where $\phi$ is
true.   (Notice that if there are no worlds where $\phi$ holds, then 
$\closest_M(w,\phi) = \emptyset$.)
\begin{itemize}
\item $(M,w) \sat \phi \RCond \psi$ if $(M,v) \sat \psi$ for all
$v \in \closest_M(w,\phi)$.
\end{itemize}

\paragraph{Axioms:}  There are a number of well-known sound and complete
axiomatizations for counterfactual logic (see, e.g., 
\cite{Burgess81,Bell89,Chellas80,Grahne,KatsunoSatoh,Lewis71,Lewis73,Lewis75}).
Here is one, based on Burgess's axiomatization, similar in spirit to the
well-known KLM properties \cite{KLM}.

\begin{list}{}{\setlength{\labelwidth}{2in}\setlength{\leftmargin}{.3in}} 
 \item[A0.] All instances of propositional tautologies in the language
 $\L^C$. 
 \item[A1.] $\phi \RCond \phi$.
 \item[A2.] $((\phi \RCond \psi_1)\land(\phi\RCond\psi_2)) \rimp
    (\phi\RCond(\psi_1\land\psi_2))$.
 \item[A3.] $((\phi_1\RCond\phi_2)\land(\phi_1\RCond\psi)) \rimp
    ((\phi_1\land\phi_2) \RCond \psi)$.
 \item[A4.] $((\phi_1\RCond\psi)\land(\phi_2\RCond\psi)) \rimp
    ((\phi_1\lor\phi_2) \RCond\psi)$.
\item[A5.] $\neg (\true \RCond \false)$.
\item[A6.] $\phi \rimp (\psi \dimp (\phi \RCond \psi))$.
\end{list}
There are three rules of inference: modus ponens, and the following two
rules:

\begin{list}{}{\setlength{\labelwidth}{2in}\setlength{\leftmargin}{.4in}} 
 \item[RA1.] From $\phi \dimp \phi'$ infer $(\phi\RCond\psi) \rimp
(\phi'\RCond\psi)$.
 \item[RA2.] From $\psi \rimp \psi'$ infer $(\phi\RCond\psi) \rimp
(\phi\RCond\psi')$.
\end{list}
A1--A4 and RA1--RA2 correspond to the KLM postulates REF (for
reflexivity), AND, CM (cautious monotonicity), OR, 
LLE (left logical equivalence), and RW (right weakening), respectively.  
A5 captures the requirement that $\W_w$ is nonempty.  Finally,
A6 is the axiom that makes $\RCond$ a counterfactual operator, and not
just a ``normality'' or ``typicality'' operator (so that ``normally
birds have wings'' becomes $\mathit{bird} \RCond \mathit{wing}$).  
Suppose that it is the case that if $\phi$ were true, then $\psi$ would be
true.  Then if $\phi$ is actually true, we would expect $\psi$ to be
true.  Moreover, if $\phi$ and $\psi$ are both true, then it seems
reasonable to assert that if $\phi$ were true, then $\psi$ would be
true.  On the other hand, it is not in general the case that if $\phi$
and $\psi$ are both true, then $\phi$'s are normally or typically
$\psi$'s. 

In his semantics, Lewis allows there to be more than one world closest
to a world $w$ where $\phi$ is true (except in the special case that
$\phi$ is true at $w$; in this case, $w$ is the unique world closest to
$w$ satisfying $\phi$).  By way of contrast, Stalnaker
\citeyear{Stalnaker68} essentially assumes that for each world $w$
and formula $\phi$, there is a unique world closest to $w$ satisfying
$\phi$.  The standard approach to getting uniqueness is to require that
$\preceq_w$ be a strict total order (whose least element is $w$); that
is, for all worlds $w' \ne w''$, either $w' \prec_w w''$ or $w'' \prec_w
w'$.  This assumption is captured by the following axiom: 
\begin{list}{}{\setlength{\labelwidth}{2in}\setlength{\leftmargin}{.3in}} 
\item[A7.] $(\phi \RCond (\psi_1 \lor \psi_2)) \rimp ((\phi \RCond
\psi_1) \lor  (\phi \RCond \psi_2))$.
\end{list}

Let $\AX$ be the axiom system consisting of axioms A0--A6 and rules of
inference RA1, RA2, and modus ponens; let $\AXp$ be $\AX$ together with
the axiom A7.  Let $\M(\Phi)$ be the collection of all counterfactual
structures over the primitive propositions in $\Phi$ (i.e.,
structures where $\pi$ interprets formulas in $\Phi$); let $\M^+(\Phi)$
be the subset of $\M(\Phi)$ consisting of all  
counterfactual structures where $\preceq_w$ is a total strict order.
As usual, I omit the $\Phi$ when it is clear from context or irrelevant.

\thm {\rm \cite{Burgess81}}\label{thm:comp} $\AX$ (resp., $\AXp$) is a
sound and 
complete axiomatization for 
the language $\Lprop$ with respect to $\M$ (resp., $\M^+$).
\ethm

\section{Relating causal models to counterfactual
structures}\label{sec:main} 
As the title suggests, in this section I take a closer look at the
relationship between causal models and counterfactual structures.
On the surface, the two approaches are quite different.  Consider the
causal model for the forest fire example, discussed  in
Section~\ref{sec:causalmodel}.  If we want to capture the forest fire
using counterfactual structures, perhaps the most natural way to do it
is to have worlds in the counterfactual structure that correspond to the
eight possible settings of the three exogenous variables ($\ML$, $L$,
and $\FF$).  We can take primitive propositions that correspond to the
settings of these variables as well; that is, the primitive propositions
have the form $\ML = i$, $L=i$, and $\FF=i$, for $i = 0,1$.  The actual
world $w$ is the one where $\ML=1 \land L=1 \land \FF=1$ holds.  The closest
world relation is described in the obvious way by the equations.  For
example, if we consider the conjunctive model, where $\FF =
\min(\ML,L)$, so that both the match and the lightning are required to
start the fire, then the closest world to $w$ where $\ML=0$ is the one
where $\ML=0 \land L=1 \land \FF=0$ holds.  On the other hand, in the
disjunctive model, where $\FF=\max(\ML,L)$, the closest world to $w$
where $\ML=0$ holds is the one where $\ML=0 \land L=1 \land \FF=1$
holds.  There is a sense in which the causal model and the corresponding
counterfactual structure(s) constructed this way satisfy the same
formulas.  In this section, I make this intuition  precise.  More
generally, I show that to every causal model in $\Trec$, there is a
corresponding counterfactual structure that satisfies the same formulas;
however, this is not the case for every causal model in $\Tun$. 

Given a signature $\S = (\U,\V,\R)$, consider the set $\Phi_\S$ of 
primitive propositions of the form $X=x$ for $X \in \V$ and $x \in
\R(x)$.  We restrict to counterfactual structures $M = (\W,R,\pi)$ for this
set of primitive propositions, where $\pi$ is such that, for each world
$w \in \W$ and variable $X \in \V$, exactly one of the formulas $X=x$ is
true.  Call such structures \emph{acceptable}.
In an acceptable counterfactual structure, a world can be associated with
an assignment 
of values to the random variables.  An acceptable structure $(\W,R,\pi)$ is
\emph{full} if, for each assignment $v$ of values to variables and all $w
\in \W$, there is
a world in $\W_w$ where $v$ is the assignment.
(I discuss the consequences of fullness shortly.)
Let $\M_a(\Phi_\S)$ consist of
all acceptable counterfactual structures over $\S$; 
let $\M_f(\Phi_\S)$ consist of
all full acceptable counterfactual structures over $\S$;  
let $\M_a^+(\Phi_\S) = \M_a(\Phi_\S) \inter \M^+(\Phi_\S)$; and let    
$\M_f^+(\Phi_\S) = \M_f(\Phi_\S) \inter \M^+(\Phi_\S)$.

As before, I identify $[\vec{Y} = \vec{y}](X=x) \in \Lprop(\S)$ with the
formula $\vec{Y} = \vec{y} \RCond (X=x) \in
\L_1^C(\Phi_{\S})$. 
I abuse notation and use $\Lprop(\S)$ to denote
the sublanguage of $\L_1^C(\Phi_{\S})$ that arises via this identification.
The following is easy to show.

\pro\label{causaltolewis} C3 and C4 are valid in $\M_a(\Phi_\S)$; C2 is
valid in $\M^+_a(\Phi_\S)$; C1 is valid in $\M_f(\phi_\S)$.  
\epro

Notice that, in $\M_a(\Phi_\S)$ (and hence all of its subsets),
the following two formula schemes are valid:
\begin{list}{}{\setlength{\labelwidth}{2in}\setlength{\leftmargin}{.3in}} 
\item[V1.] $\lor_{x \in \R(X)} X=x$
\item[V2.] If $x \ne x'$, then $X=x \rimp X \ne x'$.
\end{list}
In $\M_f(\Phi_\S)$, the following axiom is valid:
\begin{list}{}{\setlength{\labelwidth}{2in}\setlength{\leftmargin}{.3in}} 
\item[V3.] $\neg [\vec{X} = \vec{x}]\false$.
\end{list}
V3 is valid in a structure where, for all worlds $w$, there is a closest
world to $w$ such that
$\vec{X} = \vec{x}$, that is,
$\closest_M(w,\vec{X} = \vec{x}) \ne \emptyset$, which is
exactly what fullness ensures.%
\footnote{Galles and Pearl did not discuss the axioms V1--V3, but it is
clear that they are implicitly assuming that they hold.}

With this background, I can give an ``axiomatic'' proof of
Proposition~\ref{causaltolewis}:
\begin{itemize}
\item C1 follows easily from A0, A2, A5, and V2.
\item C3 is a special case of A3. 
\item C4 follows easily from A1 and RA2.
\item C2 is not provable in $\AX$ (and is not valid in $\M_a(\Phi_\S)$),
but it is provable in $\AX'$ together with the axiom V1.
Indeed, it follows easily from A1, A7, RA2, and V1 (since $\phi \rimp
\lor_{x \in \R(X)} X = x$ is valid in $\M(\Phi_\S)$).  The fact that C2
requires A7 is not surprising.   C2 is 
essentially expressing the uniqueness of solutions, which is being
captured by the assumption that there is a unique closest world where the
antecedent is true, an assumption captured by A7.  
\end{itemize}
Reversibility (C5) is conspicuously absent from this list.  
Indeed, as the following example shows, C5 is not sound in
counterfactual structures.

\xam\label{xam:C5} {\em Suppose that $\V = \{X_1,X_2,X_3\}$, and $\R(X_1) =
\R(X_2) = \R(X_3) = \{0,1\}$.  Consider a structure $(\W,R,\pi) \in
\M^+_f$ where 
there is exactly one world in $\W$ corresponding to each of the 8 possible
assignments of values to the variables.  Thus, we can identify a world 
$w$ in $\W$ with a triple $(b_1,b_2,b_3)$, where, in $w$, $X_i = b_i$.  
All that matters about $R$ is that $(0,0,0) \prec_{(0,0,0)} (1,0,0)
\prec_{(0,0,0)} (1,1,1)$, with the other 5 worlds being further from
$(0,0,0)$ than $(1,1,1)$.  Then it is immediate that 
$$(M,(0,0,0)) \sat [X_1 \gets 1; X_2 \gets 1](X_3 = 1) \land
[X_1 \gets 1; X_3 \gets 1](X_2 = 1) \land
[X_1 \gets 1](X_2 = 0).$$
Thus, C5 is violated.}
\exam

For the remainder of this section, fix $\S = (\U,\V,\R)$. 
Say that a counterfactual structure $M = (\W,R,\pi)$ in $\M^+_f(\Phi_\S)$ is
\emph{recursive} if there is a total
ordering $\prec_M$ of the variables in $\V$ such that if $W \prec_M Y$,
then for all $\vec{X} \subseteq \V-\{Y,W\}$, 
for each world $w \in \W$,
in the closest world to $w$
where $\vec{X} = \vec{x}$ and $Y=y$, the value of $W$ is the same as in
the closest world to $w$ where $\vec{X} = \vec{x}$.  Intuitively,
setting $Y$ to $y$ has no further effect on the value of $W$ once
$\vec{X}$ is set to $\vec{x}$.  Let $\Mrec$ consist of the recursive
structures in $\M_f^+$.  It is easy to see that the structure $M$
considered in Example~\ref{xam:C5} is not in $\Mrec$.  For if it
were, then we would have either $X_2 \prec_M X_3$ or $X_3
\prec_M X_2$.  If $X_3 \prec_M X_2$,
then the value of $X_3$ would have to be the same in the closest world
to $(0,0,0)$ where $X_1=1$ and $X_2=1$ as in the closest world to $(0,0,0)$ 
where $X_1 = 1$.  But it is not.  The same
problem occurs if $X_2 \prec_M X_3$.  

\pro\label{pro:recursive} C5 is valid in $\Mrec$. \epro

\prf Suppose that $M = (\W,R,\pi) \in \Mrec$, $w \in \W$, and the
antecedent of C5 holds at $(M,w)$.  If $Y \prec_M W$, then 
it is immediate from the fact that $(M,w) \sat [\vec{X} \gets \vec{x}; Y
\gets y](W = w)$ that we also have $(M,w) \sat [\vec{X} \gets
\vec{x}](W=w)$.  Now suppose that $W \prec_M Y$.  Then from 
$(M,w) \sat [\vec{X} \gets \vec{x}; W \gets w](Y = y)$ it follows that
$(M,w) \sat [\vec{X} \gets \vec{x}](Y = y)$.  Suppose, by way of
contradiction, that 
$(M,w) \sat [\vec{X} \gets \vec{x}](W = w')$ for some $w' \ne w$.  
By C3 (which, by Proposition~\ref{causaltolewis}, is sound in $M$), it
follows that  
$(M,w) \sat [\vec{X} \gets \vec{x}; Y = y](W=w')$.  This, combined with
the fact that $(M,w) \sat [\vec{X} \gets \vec{x}; Y = y](W=w)$,
contradicts C1 (which, by Proposition~\ref{causaltolewis}, is also sound
in $M$).  (Note that this argument actually shows that the stronger
version of C3 used by Galles and Pearl, discussed in
Footnote 4, follows from the weaker version used here.)\ 
\eprf

Not surprisingly, the proof of Proposition~\ref{pro:recursive} is
essentially identical to the argument given by Galles and Pearl that
reversibility holds in recursive structures.  Indeed, it is not hard to
show 
that the axiom C6 that characterizes recursive structures is valid in
$\Mrec$, and C5 follows from the other axioms in the presence of C6.

Even more can be shown.  In a precise sense, every causal model
in $\Trec(\S)$ can be identified with a counterfactual structure in
$\Mrec(\S)$ where the same formulas are true.  It follows that recursive
counterfactual structures are at least as general as recursive causal
models.  The converse is also true.  I now make these claims precise.

Given a causal model $T = (\S,\F) \in \Trec$, we construct a model $M_T =
(\W,R,\pi)$ as follows.  Let $\W$
consist of all the assignments of values to the variables in $\U \union
\V$.  The interpretation $\pi$ is defined in the obvious way: $X=x$ is
true in $w$ if $w$ assigns $X$ value $x$.  For each context $\vec{u}$, 
let $\vec{v}_{\vec{u}}$ be the assignment to the
variables in $\V$ that is forced by the equations.  (Since $T \in
\Trec$, $\vec{v}_{\vec{u}}$ is uniquely determined.)  More generally, for
each assignment $\vec{X} \gets \vec{x}$, let 
$\vec{v}_{\vec{u},\vec{X} \gets \vec{x}}$ be the assignment to the
variables in $V$ determined by the equations $\F^{\vec{X} \gets
\vec{x}}$ in the context $\vec{u}$.  Let
$w_{\vec{u}} = (\vec{u},\vec{v}_{\vec{u}})$ and let $w_{\vec{u},\vec{X}
\gets \vec{x}} = (\vec{u}, \vec{v}_{\vec{u},\vec{X} \gets \vec{x}})$.
Finally, let $R$ be such that the closest world to $w_{\vec{u}}$ where
$\vec{X} = \vec{x}$ is $w_{u,\vec{X} \gets \vec{x}}$, and for all
assignments $\vec{u}$ and $\vec{v}$ to the exogenous and endogenous
variables, respectively, $\W_{\vec{u},\vec{v}}$ consists of all worlds
$(\vec{u},\vec{v}')$ such that $\vec{v}'$ is an arbitrary assignment to the
endogenous variables.  This does not
uniquely determine $R$.  Indeed, it places no constraints on $\preceq_w$
if $w$ is not of the form $w_{\vec{u}}$ and does not completely
determine $R$ even if $w$ does have the form $w_{\vec{u}}$.  It is easy
to define a relation $R$ that satisfies these constraints such that 
(1) $\prec_{w_{\vec{u}}}$ is a strict total order on $\W_{w_{\vec{u}}}$
and (2) for  $w$ not of the form $w_{\vec{u}}$, $\prec_w$ is a strict total
order on $\W_w$ that
satisfies the recursiveness constraint.
\shortv{The following theorem, whose proof is left to the full paper, is
now easily shown.}

\thm\label{thm:map1} $M_T \in \Mrec$ and, for all formulas $\phi \in
\Lprop(\S)$, we have $(T,\vec{u}) \sat \phi$ iff $(M_T,w_{\vec{u}}) \sat
\phi$.  \ethm

\fullv{\prf It is easy to see that we can take $\prec_{M_T} = \prec_T$, so $M_T
\in \Mrec$.  Using the definition of $R$, it is easy to show by
induction on the structure of formulas that $(T,\vec{u}) \sat \phi$ iff
$(M_T,w_{\vec{u}}) \sat \phi$.  I leave the details to the reader. \eprf}

Theorem~\ref{thm:map1} shows that we can embed $\Trec$ in $\Mrec$.  I
next give an embedding of $\Mrec$ in $\Trec$.  Now the causal model in
$\Trec$ depends both on the counterfactual structure and a world in that
structure.  (I discuss why this is so after Theorem~\ref{thm:map2}.)
Suppose that $M = (\W,R,\pi) \in \Mrec(\S)$ and $w \in \W$.
Suppose that $\V = \{X_1, \ldots, X_n\}$ and,
without loss of generality, that $X_i \prec_M X_j$ iff $i < j$.
Consider the causal model $T_{M,w} = (\S,\F)$, where $F_X$ is defined by
induction on the $\prec_M$-ordering.  That is, we first define $F_{X_1}$,
since $X_1$ is the $\prec$-minimal variable, then define $F_{X_2}$, and so
on.  Suppose that $(M,w) \sat X_1 = x_1$.  Then define
$F_{X_1}$ so that, for all contexts $\vec{u}$ and assignments
$\vec{v}$ to the variables in $\V - \{X_1\}$, $F_{X_1}(\vec{u},\vec{v})
= x_1$.  Suppose that we have defined $F_{X_i}$ for $i \le k$.  Define
$F_{X_{k+1}}$ so that, for all contexts $\vec{u}$ and
all assignments 
$\vec{v}$ to the variables in $\V - \{X_{k+1}\}$, we have 
$F_{X_{k+1}}(\vec{u},\vec{v}) = x$ iff $(M,w) \sat [X_1 \gets v_1;
\ldots, X_k \gets v_k](X_{k+1} = x)$.  
\shortv{Again I leave the straightforward proof of the following theorem
to the full paper.}

\thm\label{thm:map2} 
$T_{M,w} \in \Trec$ and, for all formulas $\phi \in
\Lprop(\S)$ and all contexts $\vec{u}$, we have $(M,w) \sat
\phi$ iff $(T_{M,w},\vec{u}) \sat \phi$.
\ethm

\fullv{\prf It is easy to see that $T_{M,w} \in \Trec$; the definition of $\F$
guarantees that $\prec_{T_{M,w}} = \prec_M$.  The definition of $F_X$ is
independent of the context; it easily
follows that for all contexts $\vec{u}$ and $\vec{u}'$,
we have $(T_{M,w},\vec{u}) \sat \phi$ iff $(T_{M,w},\vec{u}')
\sat \phi$.  Again, an easy induction on the structure of $\phi$ shows that,
for all contexts $\vec{u}$,  $(M,w) \sat \phi$ iff $(T_{M,w},\vec{u}) \sat
\phi$. \eprf }

It is easy to modify the construction slightly so that each context
$\vec{u}$ corresponds to a different world $w \in \W$.  Thus, if the
number of contexts is at least $|\W|$, then we can get a closer analogue
to Theorem~\ref{thm:map1}, where we can associate with each world $w \in
\W$ a context $\vec{u}_w$ and show that $(M,w) \sat \phi$ iff
$(T,\vec{u}_w) \sat \phi$.  However, Theorem~\ref{thm:map2} suffices for
the following corollary.

\cor\label{cor:equiv} The same formulas in $\Lprop(\Phi_\S)$ are valid in both 
$\Trec(\S)$ and $\Mrec(\S)$.\ecor

\prf If $\phi$ is not valid in $\Trec$, then there is some causal model $T
\in \Trec$ and context $\vec{u}$ such that $(T, \vec{u}) \sat \neg
\phi$.  By Theorem~\ref{thm:map1}, $(M_T,w_{\vec{u}}) \sat \neg \phi$,
so $\phi$ is not valid in $\Mrec$.  For the converse, if $\phi$ is not
valid in $\Mrec$, then there is some counterfactual structure $M \in
\Mrec$ and 
world $w$ such that $(M,w) \sat \neg \phi$.  By Theorem~\ref{thm:map2},
$(T_{M,w}, \vec{u}) \sat \neg \phi$.  Thus, the same formulas are
satisfiable in both $\Trec$ and $\Mrec$, and hence the same formulas are
valid. \eprf

What happens if we consider $\Tun$ rather than $\Trec$?  Now causal
models and counterfactual structures are incomparable.   Consider the
formula 
$$\begin{array}{ll}
\phi \eqdef 
&[X_1 \gets 1](X_2=1 \land X_3=0) \, \land\\ 
&[X_2 \gets 1](X_3=1 \land  X_1=0) \, \land\\ 
  &[X_3\gets 1](X_1=1 \land X_2=0).
\end{array}$$

\thm\label{thm:counterexample} $\neg \phi$ is valid in $\Trec$ and
$\M_f$ (and hence $\M_f^+$), but $\phi$ is satisfiable in
$\Tun$. \ethm 

\prf 
The validity of $\neg \phi$ in $\Trec$ follows from its validity in
$\M_f$, in light of Corollary~\ref{cor:equiv} and the fact that 
$\Mrec \subseteq \M_f$.  Nevertheless, I prove the validity of $\neg \phi$
in $\Trec$ first, since the proof is short and gives some insight.
Consider a
causal model $T \in \Trec$, and suppose, by way of contradiction, that
$(M,w) \sat \phi$.  Let $\prec_T$ be the total ordering on the
variables in $\V$ in $T$.  
One of $X_1$, $X_2$, and $X_3$ must be minimal with respect to
$\prec_T$.  Suppose it is $X_1$ and that $(T,\vec{u}) \sat X_1 = i$.
Then we must have $(T,\vec{u}) \sat [X_2 = 1](X_1 = i)$ and 
$(T,\vec{u}) \sat [X_3 = 1](X_1 = i)$.  But since $(T,\vec{u}) \sat
\phi$, it follows that $(T,\vec{u}) \sat [X_2=1](X_1 = 0)$ and 
$(T,\vec{u}) \sat [X_3=1](X_1 = 1)$.  Thus, $X_1$ cannot be minimal with
respect to $\prec_T$.  An analogous argument shows that $X_2$ and $X_3$
also cannot be minimal with respect to $\prec_T$.  Thus, we have a
contradiction.  

I next show that $\neg \phi$ is valid in
$\M_f$.  Suppose by way of contradiction that $M \in \M_f$ and $(M,w)
\sat \phi$.  
Consider a world $w'$ closest to $w$ that satisfies $X_1 = 1
\lor X_2 =1 \lor X_3 = 1$.  
(Since $M \in \M_f$, there is guaranteed to be such a world.)
Suppose that $(M,w') \sat X_1 = 1$.  Note
that $w'$ must be one of the worlds closest to $w$ that satisfies $X_1
= 1$.  Since $(M,w) \sat [X_1 = 1](X_2 =1)$, we must have $(M,w') \sat
X_2 =1$.  Thus, $w'$ must also be one of the worlds closest to $w$ that
satisfies $X_2 = 1$.  Since $(M,w) \sat [X_2 = 1](X_3 = 1)$, we must
have that $(M,w') \sat X_3 = 1$.  On the other hand, since $(M,w) \sat
[X_1 = 1](X_3 = 0)$, we must have have that $(M,w') \sat X_3 = 0$.  This
gives a contradiction.  
A similar contradiction
arises if $(M,w') \sat X_2 = 1$ or if $(M,w') \sat X_3 = 1$.  Since 
$(M,w') \sat X_1 = 1 \lor X_2 = 1 \lor X_3 = 1$ by construction, this
gives a contradiction to the assumption that $(M,w) \sat \phi$.  
\commentout{
Purely syntactic proof: we use the fact that 
$[A \union B](B \union C) \land [B]D \rimp [A \union B]$.  This can be
proved as follows: $[A \union \neg B](B \rimp D) (follows easily from A1
and RA2); $[B]D \rimp [B](B \rimp D)$ (again, using RA2, since $D \rimp
(B \rimp D)$ is valid).  Thus, by A4, $[A \land \neg B \union B](B \rimp D)$.
The desired result now follows from RA1.  Also use the OR rule to show
that  $[X_1 = 1 \lor X_2 = 1 \lor X_3 = 1](X_1 = 0 \lor X_2 = 0 \lor X_3
= 0)$.  
}

Finally, I must show that there is a causal model $T \in \Tun$ and a
context $\vec{u}$ such that $(T,\vec{u}) \sat \phi$.  Let $\V =
\{X_1,X_2,X_3\}$ and 
let $\U = \{U\}$.  Take $\R(X_1) = \R(X_2) = \R(X_3) = \{0,1\}$ and
$\R(U) = \{0\}$.  In defining $\F$, I write $F_i$ instead of $F_{X_i}$,
and omit the $U$ argument (since it is always 0).  Thus, $F_1(0,0) = 0$
means that when $X_2 = X_3 = 0$, then $X_1 = 0$.  Define
\begin{itemize}
\item \mbox{$F_1(0,0) = 0$;
\journal{\item} $F_1(0,1) = 1$;
\journal{\item} $F_1(1,0) = 0$;
\journal{\item} $F_1(1,1) = 0$;}
\item \mbox{$F_2(0,0) = 0$;
\journal{\item} $F_2(0,1) = 0$;
\journal{\item} $F_2(1,0) = 1$;
\journal{\item} $F_2(1,1) = 0$;}
\item \mbox{$F_3(0,0) = 0$;
\journal{\item} $F_3(0,1) = 1$;
\journal{\item} $F_3(1,0) = 0$;
\journal{\item} $F_3(1,1) = 0$.}
\end{itemize}

I must now verify that $T = (\S,\F) \in \Tun$, and that $(T,0) \sat
\phi$.  This is straightforward, although tedious.
First observe that $(0,0,0)$ is the only
solution to all the equations in the basic causal model.  It is easy to
see that $(0,0,0)$ is a solution.  To see that it is the only solution,
observe that $(0,0,1)$ cannot be a solution because $F_3(0,0) =
0$; similarly, $(0,1,0)$ and 
$(1,0,0)$ cannot be solutions; $(0,1,1)$ cannot be a
solution because $F_2(0,1) = 0$; $(1,1,0)$ cannot be a solution because  
$F_1(1,0) = 0$; $(1,0,1)$ cannot be a solution because $F_3(1,0) = 0$; finally,
$(1,1,1)$ cannot be a solution because $F_1(1,1) = 0$.  

It is clear that there must be a unique solution if we set two of the
three variables (forced by the equation for the third variable); for
example, if $X_1\gets 1$ and $X_2 \gets 0$, then the unique solution is
$(1,0,0)$.  
If $X_1$, $X_2$, or $X_3$ is set to 0, then one solution is $(0,0,0)$. 
The solution is unique for the same reasons that $(0,0,0)$ was the
unique solution to the original collection of equations.
Finally, if $X_1 \gets 1$, then
$(1,1,0)$ is a solution; if $X_2 \gets 1$, then $(0,1,1)$ is a solution;
and if $X_3 \gets 1$, then 
$(1,0,1)$ is a solution.  We must show that there are no other solutions
in all three cases.  If $X_1\gets 1$, (1,0,0) is not a solution since $F_2(1,0) =
1$; $(1,0,1)$ is not a solution because $F_3(1,0)=0$; and $(1,1,1)$ is not a
solution because $F_3(1,1) = 0$.  If $X_2 \gets 1$, $(0,1,0)$ is not a
solution because $F_3(0,1) = 1$; $(1,1,0)$ is not a solution because
$F_1(1,0) = 0$; $(1,1,1)$ is not a solution because $F_1(1,1) = 0$.
Finally, if $X_3 \gets 1$, 
$(0,0,1)$ is not a solution because $F_1(0,1) = 1$; $(0,1,1)$ is not a
solution  because $F_2(0,1) = 0$; and $(1,1,1)$ is not a solution
because $F_1(1,1) = 0$.  Thus, $T \in \Tun$.

It is now straightforward to check that $(T,u) \sat \phi$.
\eprf

In Section~\ref{sec:causalmodel}, I argued that the choice of language was
significant.  All the results of this section are stated for the
language $\Lprop$.  In light of Lemma~\ref{lem:equiv},
Theorems~\ref{thm:map1} and \ref{thm:map2} apply without change to the 
language  $\Lex$.  Moreover, since $\Tun \subseteq \T$ and $\Lprop \subseteq
\Lex$, the formula $\neg \phi$ from Theorem~\ref{thm:counterexample}
(which is in $\Lprop$, and hence 
also in $\Lex$) continues not to be valid in $\T$, while being valid
in counterfactual structures.  While  C5 is not valid in $\T$, the
generalization of C5 mentioned at the end of
Section~\ref{sec:causalmodel} is valid in $\T$ \cite{Hal20}, and is not
valid in counterfactual structures.  So the full class $\T$ of
causal models is incomparable in expressive power to counterfactual
structures with respect the language $\Lex$ (which is the language
arguably most appropriate for $\T$).   
Formally, we have:
\cor $\T$ is incomparable in expressive power to $\M$ with respect to
the language $\Lex$.
\ecor

\section{Discussion}\label{sec:discussion}
I have shown that 
the expressive power of causal models as models for counterfactuals
is incomparable to that of the Lewis-Stalnaker ``closest-world'' 
possible-worlds semantics for counterfactuals; thus, the definition of
counterfactuals in causal models is \emph{not}, in general, compatible
with the axioms of possible world semantics, although it is if we
restrict to recursive causal models.   

Specifically, causal models
where the equations are recursive can be viewed as defining a strict
subclass of the standard possible-worlds semantics.   
More precisely, a set 
of structural equation defines a world $w$ (characterized by the unique
solution to the equations) and can be implicitly viewed as defining an
ordering relation on worlds such that, for every formula $\vec{X} =
\vec{x}$, the solution to the equations when $\vec{X}$ is set to
$\vec{x}$ determines a world $w_{\vec{X} = \vec{x}}$ that is 
the world closest to $w$ according to the ordering such that $\vec{X}
= \vec{x}$. 
Somewhat surprisingly, this is not the case if we go to the larger class
of causal models defined by equations that are not recursive, but
have a unique solution for all settings $\vec{X} = \vec{x}$.  Of course,
it is still the case that there is a world $w_{\vec{X} = \vec{x}}$
determined by the equations when $\vec{X}$ is set to $\vec{x}$.
However, there 
is, in general, no ordering on worlds such that $w_{\vec{X} = \vec{x}}$
is the closest world to $w$ according to the ordering.  A closeness
ordering on worlds places some restrictions (e.g., those characterized
by the formula $\neg \phi$ in Theorem~\ref{thm:counterexample}) that do
not hold in all causal models in $\Tun$.

So where does this leave us?  It is still the case that, in causal
models, a formula such as $\phi \RCond \psi$ is true at a world $w$
if $\psi$ is true at some appropriate world $w'$ satisfying $\phi$.
However, $w'$ cannot 
be viewed as the ``closest'' world to $w$ satisfying $\phi$.  This
leaves open the question of whether there are other ways of defining
``appropriateness'' other than ``closeness''.  I do not have strong
intuitions here, but it is a question that is perhaps worth pursuing. 
My own feeling is that these arguments show that models in $\Tun -
\Trec$ are actually not good models for causality.  It is quite
difficult to verify that a nonrecursive causal model is in $\Tun$, as
the model $T$ given in Theorem~\ref{thm:counterexample} satisfying $\phi$
shows.  I am not aware of any interesting real-world situation that is
captured by a model in $\Tun-\Trec$.  Further evidence of the
``unreasonableness'' of models in $\Tun-\Trec$ is given by recent work
of Zhang, Lam, and de Clerq \citeyear{ZYC12}.  They show that although the
reversibility axiom blocks cycles of counterfactual dependence of length
two, it does not block longer cycles.  Indeed, they observe that the
causal model $T$ of Theorem~\ref{thm:counterexample} has a cycle of
length three.  

Does this mean that we should restrict to recursive models?   There are
certainly equations in physics (e.g., those connecting pressure and
volume) that exhibit circular dependencies.  Perhaps nonrecursive models
would be 
appropriate for them (although once we add time to the picture, we
may well be able to use a recursive model to capture any
particular scenario).  In a general nonrecursive model, there may be
several solutions to an intervention (see \cite[Appendix A.4]{HP01b} for
further discussion of this point).  But this just corresponds to there
being several worlds satisfying a formula $\phi$ that are closest to a
given world, which is certainly allowed in Lewis's framework.  I suspect
that there is an interesting class of nonrecursive causal models that
can be captured in Lewis's framework, and that these will turn out to be
the models that actually arise in practice.  

Zhang \citeyear{Zhang12} makes some progress on this issue.
He proposes two condition on causal models, which he calls
\emph{solution-ful} and \emph{solution-conservativeness}.
The former condition is easy to understand: a causal model $T$ is
solution-ful if, for every context $\vec{u}$, the equations have a
solution (not necessarily unique).  The second condition is somewhat
more  complicated.  $T$ is solution-conservative if, for every context
$\vec{u}$, if a solution to $T_{\vec{X} \gets \vec{x}}$ is consistent
with $\vec{Y}=\vec{y}$, then every solution to $T_{\vec{X} \gets \vec{x}
\land \vec{Y} = \vec{y}}$ is a solution to $T_{\vec{X} \gets \vec{x}}$.
Zhang shows that a causal model satisfies  
these two conditions iff it satisfies analogues of all the axioms in one
of Lewis's axiomatizations of causal counterfactuals
\cite[p.~973]{Lewis73}.  Thus, these two conditions are necessary for a
causal model to be translatable to a counterfactual  structure.  They
are not sufficient, since they are satisfied by all models in $\Tun$.
There is clearly more to be done in understanding the connections between
causal models and counterfactual structures.

Turning to a more technical point, it is worth trying to
understand in more detail exactly what goes wrong in the Galles-Pearl
argument.  When making the comparison, Galles and Pearl did not use the
axiom system $\AX$.  Rather, they used one of Lewis' axiomatizations of
conditional logic from \cite{Lewis73}.  Nevertheless, the problem with
their argument can be understood in terms of $\AX$.  Galles and Pearl
show that all but one of the axioms in the system they consider follows
from C3, C4, and C5.  (They actually also implicitly use C1 and C2, but
this is a minor point.)  The remaining axiom is \journal{$$((\phi_1 \lor
\phi_2) \RCond \phi_1) \lor ((\phi_1 \lor 
\phi_2) \RCond \phi_2) \lor [((\phi_1 \lor \phi_2) \RCond \psi) \dimp
((\phi_1 \RCond \psi)\lor (\phi_2 \RCond \psi))].$$}
\fullv{$$\begin{array}{ll}
((\phi_1 \lor \phi_2) \RCond \phi_1) \lor ((\phi_1 \lor
\phi_2) \RCond \phi_2) \, \lor \\
{[((\phi_1 \lor \phi_2) \RCond \psi) \dimp
((\phi_1 \RCond \psi)\lor (\phi_2 \RCond \psi))]}.
\end{array}
$$}
For this axiom, they say ``Because actions in causal models are
restricted to conjunctions of literals that is, in the language of this
paper, because in a formula in $\Lprop$ of the form $\phi \RCond \psi$,
$\phi$ is a conjunction of formulas of the form $X=x$], this axiom is
irrelevant.''  They thus ignore the axiom.  Unfortunately, this argument
is flawed.  If it were true, then it would be the case that the class of
causal models in $\Tun$ would be \emph{less} general than counterfactual
structures since causal models would satisfy more axioms---all the
relevant axioms satisfied by counterfactual structures, and, in
addition, C5 (reversibility).  
However, as
we have seen, the formula $\phi$ of Theorem~\ref{thm:counterexample} is
satisfiable in $\Tun$, while $\neg \phi$ is valid in $M_f$.  The formula 
$\neg \phi$ is equivalent to
\journal{$$([X_1 \gets 1](X_2=1 \land X_3=0) \land 
[X_2 \gets 1](X_3=1 \land  X_1=0) \rimp \neg[X_3\gets 1](X_1=1 \land
X_2=0).$$}
\fullv{$$\begin{array}{ll}
([X_1 \gets 1](X_2=1 \land X_3=0) \land 
[X_2 \gets 1](X_3=1 \land  X_1=0)\\
\ \ \  \rimp \neg[X_3\gets 1](X_1=1 \land
X_2=0).\end{array}$$} 
Thus, $\neg \phi$ is a formula whose antecedent and conclusion are both
formulas in $\Lprop$.   
The argument for the validity of $\neg \phi$ given in the proof of
Theorem~\ref{thm:counterexample} is purely semantic, but, as
I show in the appendix, $\neg \phi$ can
also be derived from $\AX$ 
together with V2 and V3.  The derivation
uses A4 in a crucial way.  
Note that A4 
has disjunctions on the left-hand side of $\RCond$ and thus
cannot be expressed in $\Lprop$.  But it is
not irrelevant!   
Indeed, an easy argument (given in the
appendix) shows that $\neg 
\phi$ cannot be derived using $\AX - \{\mbox{A4}\}$ together with V2 and V3 if
we restrict to formulas in $\Lprop$.  
Thus, we can start with 
assumptions in the language $\Lprop$, 
end up with a conclusion in $\Lprop$, but have a derivation that, along the
way, uses A4 and has steps that involve formulas with disjunctions on
the left-hand side of $\RCond$.  The Galles and Pearl argument ignores
this possibility.

\commentout{
That axiomatization used A0 and
A1 above, modus ponens, 
and the following additional axioms and rule (using a numbering that
essentially corresponds to that used by Galles and Pearl):
\begin{itemize}
\item[D3.] $((\phi_1 \RCond \phi_2) \land (\phi_2 \RCond \phi_1)) \rimp
((\phi_1 \RCond \psi) \dimp (\phi_2 \RCond \psi))$
\item[D4.] $((\phi_1 \lor \phi_2) \RCond \phi_1) \lor ((\phi_1 \lor
\phi_2) \RCond \phi_2) \lor [((\phi_1 \lor \phi_2) \RCond \psi) \dimp
((\phi_1 \RCond \psi)\lor (\phi_2 \RCond \psi))]$
\item[D5.] $(\phi \RCond \phi) \rimp (\phi \rimp \psi)$
\item[D6.] $(\phi \land \psi) \rimp (\phi \RCond \psi)$
\item[RD1.] From $\phi_1 \land \ldots \land \phi_n) \rimp \psi$ infer
$(\sigma \RCond \phi_1) \land \ldots \land (\sigma \RCond \phi_n)) \rimp
(\sigma \rimp \psi)$.
\end{itemize}

D5 and D6 are together easily seen to be equivalent to A6, RC1 follows
easily from A2 and RA2, and it is not hard to show that D3 follows from
$\AX$.  Since it will be useful later, I sketch the argument here.
By symmetry, it suffices to show  that $((\phi_1 \RCond \phi_2) \land (\phi_2
\RCond \phi_1)) \land (\phi_1 \RCond \psi)) \rimp (\phi_2 \RCond \psi)$
is provable.  By A3, we can conclude $\phi_1 \land \phi_2 \RCond \psi$
from the antecedents.  Using RA2, we can get 
\begin{equation}\label{eq1}
(\phi_1 \land \phi_2) \RCond \psi \lor \neg \phi_1
\end{equation}
from the antecedents.  By A1 and RA2, it follows that
\begin{equation}\label{eq2}
(\neg \phi_1 \land \phi_2) \RCond  \psi \lor \neg \phi_1
\end{equation}
is provable.
Thus, applying A4 and RA1 to (\ref{eq1}) and (\ref{eq2}), we can
conclude $\phi_2 \RCond \psi \lor \neg \phi_1$ from the antecedents.
Since we also have $\phi_2 \RCond \phi_1$ in the 
antecedent, by A2 and RA2, we can conclude $\phi_2 \RCond \psi$ from the
antecedents, as desired.
Note that A4 (the OR rule in the KLM system) plays a significant role
in this argument. 

What about D4?  It is trivially provable in $\AXp$; indeed, the stronger
formula consisting of just the first two disjuncts, 
$((\phi_1 \lor \phi_2) \RCond \phi_1) \lor ((\phi_1 \lor
\phi_2) \RCond \phi_2)$ follows easily from A1 and A7.  On the other
hand, D4 is not provable in $\AX$. Indeed, while the 
implication from right to left follows from the OR
rule (A4), the implication from left to right is not
valid in all models in $\M$.  (See the appendix for a counterexample.)
\fullv{For example, consider a structure $M =
(\W,R,\pi)$, where $\W = \{w_0, \ldots, w_4\}$, $\phi_1$, $\phi_2$, and
$\psi$ are primitive propositions, and $\pi$ is such that $(M,w_0) \sat
\neg \phi_1 \land \neg 
\phi_2 \land \neg \psi$, $(M,w_1) \sat \phi_1 \land \neg \phi_2 \land \psi$,  
$(M,w_2) \sat \neg \phi_1 \land \phi_2 \land \psi$,  $(M,w_3) \sat 
\phi_1 \land \neg \phi_2 \land \neg \psi$, and 
$(M,w_4) \sat \neg \phi_1 \land \phi_2 \land \neg \psi$, and $R$ is such 
that 
$w_0 \preceq_{w_0} w_1 \preceq_{w_0} w_3$ and 
$w_0 \preceq_{w_0} w_2 \preceq_{w_0} w_4$; that is, $\preceq_{w_0}$ is
characterized by the following diagram:
\begin{verbatim}
         w_1 -- w_3
       /
 w_0 /
     \ 
       \
         w_2 -- w_4
\end{verbatim}
Thus, the closest worlds to $w_0$ where $\phi \lor \phi_2$ holds are
$w_1$ and $w_2$.  Since $\psi$ is true at both $w_1$ and $w_2$, but neither
$\phi_1$ nor $\phi_2$ are true at both worlds, we must have 
$$(M,w_0) \sat \neg((\phi_1 \lor \phi_2) \RCond \phi_2) \land 
\neg((\phi_1 \lor \phi_2) \RCond \phi_2) \land 
((\phi_1 \lor \phi_2) \RCond \psi).$$
However, the closest worlds to $w_0$ where $\phi_1$ holds are 
$w_1$ and $w_3$.  Since $\neg \psi$ holds at $w_3$, we must have 
$(M,w_0) \sat \neg(\phi_1 \RCond \psi)$.  Similarly, $(M,w_0) \sat 
\neg(\phi_2 \RCond \psi)$.
}
As a consequence of this, I do not believe that Lewis's axioms are both
sound and complete for either $\M$ or $\M^+$.  This is not a significant
point.  Lewis does not provide a formal completeness proof, but 
he provides other systems that are sound and complete.  In any case, my
criticism of Galles and Pearl is independent of this issue.

When comparing Lewis's axioms to their axioms (essentially, C3, C4, and
C5), Galles and Pearl make the following points (see
\cite[Section 7.4.2]{pearl:2k}:
\begin{enumerate}
\item D3 is a weaker form of reversibility (A5)
\item Because actions in causal models are restricted to conjunctions of
literals, D4 is irrelevant in the causal model framework.
\item D5 and D6 follow directly from composition (A3).
\end{enumerate}
I do not understand the first point.  It seems to suggest that D3
follows from A5.  This is not the case.  
}

\paragraph{Acknowledgments:}  Thanks to Franz Huber, Judea Pearl, Jiji
Zhang, and the anonymous reviewers of the paper for useful comments. 

\appendix
\section{A derivation of $\neg \phi$}

In this appendix, I show that the formula
$$([X_1 \gets 1](X_2=1 \land X_3=0) \land 
[X_2 \gets 1](X_3=1 \land  X_1=0) \rimp \neg[X_3\gets 1](X_1=1 \land
X_2=0)$$
can be derived from AX together with V2 and V3.
To simplify notation, I write $\vdash \phi'$ if the formula $\phi'$ can
be derived from AX, and $\vdashp \phi'$ if $\phi'$ can be derived from
AX + $\{$V2, V3$\}$. 

I first need a technical lemma.
\lem\label{lem:technical} $\vdash (\phi_1 \RCond \phi_2  \land \phi_2
\RCond \phi_3) \rimp (\phi_1 \lor \phi_2) \RCond \phi_3$.  \elem

\prf By A1, $\vdash \phi_2 \RCond \phi_2$, and by A4, 
$\phi_1 \RCond \phi_2 \land \phi_2 \RCond \phi_2 \rimp 
(\phi_1 \lor \phi_2) \RCond \psi$.  Thus,
\begin{equation}\label{eq1}
\vdash (\phi_1 \RCond \phi_2) \rimp (\phi_1 \lor \phi_2) \RCond \phi_2.
\end{equation}

Next observe that it easily follows from A1 and A4 that 
(a) $\vdash (\phi_1 \land \neg \phi_2) \RCond (\phi_2 \rimp \phi_3)$
(since $(\phi_1 \land \neg \phi_2) \rimp (\phi_2 \rimp \phi_3)$ is a
tautology) and (b) $\vdash (\phi_2 \RCond \phi_3) \rimp 
\phi_2 \RCond (\phi_2 \rimp \phi_3)$ (since $\phi_3 \rimp (\phi_2 \rimp
\phi_3)$ is a tautology).  By A4, RA1, and the observation that 
$((\phi_1 \land \neg \phi_2) \lor \phi_2)  \dimp (\phi_1 \lor \phi_2)$
is a tautology,
we can conclude that 
\begin{equation}\label{eq2}
\vdash (\phi_2 \RCond \phi_3) \rimp ((\phi_1 \lor \phi_2) \RCond
(\phi_2 \rimp \phi_3)).
\end{equation}

Let $\psi$ be the formula $(\phi_1 \RCond \phi_2  \land \phi_2
\RCond \phi_3)$.  
From (\ref{eq1}) and (\ref{eq2}), it follows that 
$$\vdash \psi \rimp ((\phi_1 \lor \phi_2) \RCond \phi_2) \land  
(\phi_1 \lor \phi_2) \RCond (\phi_2 \rimp \phi_3)).$$
Applying A2 and RA1, it follows that 
$$\vdash \psi \rimp ((\phi_1 \lor \phi_2) \RCond \phi_3,$$
as desired.
\eprf

Now the proof is easy.  By Lemma~\ref{lem:technical}, it follows that 
\journal{
$$\vdash (X_1=1 \RCond X_2=1) \land (X_2=1 \RCond X_3 = 1) \rimp (X_1 =
1 \lor X_2 = 1) \RCond X_3 = 1.$$}
\fullv{
$$\begin{array}{ll}
\vdash (X_1=1 \RCond X_2=1) \land (X_2=1 \RCond X_3 = 1) \rimp\\
\ \ \  (X_1 =
1 \lor X_2 = 1) \RCond X_3 = 1.\end{array}$$}
Applying the lemma again, we get that 
\journal{$$\vdash (X_1 = 1 \lor X_2 = 1) \RCond X_3 = 1 \land (X_3 = 1
\RCond X_1 = 0) \rimp (X_1 = 1 \lor X_2 = 1 \lor X_3 = 1) \RCond X_1 = 0.$$}
\fullv{$$\begin{array}{ll}
\vdash (X_1 = 1 \lor X_2 = 1) \RCond X_3 = 1 \land (X_3 = 1
\RCond X_1 = 0) \rimp\\ \ \ \  (X_1 = 1 \lor X_2 = 1 \lor X_3 = 1)
\RCond X_1 = 0.\end{array}$$} 
Let $\psi'$ be an abbreviation for $X_1 = 1 \lor X_2 = 1 \lor X_3 = 1$.
Thus, $\vdash \phi \rimp (\psi' \RCond X_1 = 0)$.  An analogous argument
shows that  
$\vdash \phi \rimp \psi' \RCond X_2 = 0)$ and $\vdash \phi \rimp (\psi'
\RCond X_3 = 0)$.  By A2, we have that 
\begin{equation}\label{eq3}
\vdash \phi \rimp (\psi' \RCond (X_1 = 0 \land X_2 = 0 \land X_3 = 0).
\end{equation}
By V2, we have that $\vdashp \psi' \rimp (X_1 \ne 0 \lor X_2 \ne 0 \lor
X_3 \ne 
0)$.  Thus, by A1 and RA2, we have that 
\begin{equation}\label{eq4}
\vdashp \phi \rimp (\psi' \RCond (X_1 \ne 0 \land X_2 \ne 0 \land X_3 \ne 0).
\end{equation}
By A2, RA1, (\ref{eq3}), and (\ref{eq4}), we have that
$$\vdashp \phi \rimp (\psi' \RCond \false).$$
By V3, it follows that $\vdashp \neg \phi$, as desired.
\eprf

Note that all the axioms in AX other than A4 can be expressed using the
language $\Lprop$ (provided that the formulas $\phi$ and $\psi$
mentioned in these axioms are taken to be conjunctions of 
primitive propositions or just single primitive propositions, as
appropriate).  It is easy to check that all these axioms are valid in
$\Tun$, as are V2 and V3.  Since $\phi$ is satisfiable in $\Tun$, $\neg
\phi$ cannot be proved from $(\AX - \{\mbox{A4}\}) \union \{\mbox{V2,V3}\}$.


\bibliographystyle{chicagor}

\end{document}